\documentclass[10pt,twocolumn,letterpaper]{article}

\usepackage{cvpr}              

\usepackage{graphicx}
\usepackage{amsmath}
\usepackage{amssymb}
\usepackage{booktabs}
\usepackage{makecell}
\usepackage{bbding}
\usepackage{threeparttable}
\usepackage{adjustbox}
\usepackage{graphicx}
\usepackage{lipsum}
\usepackage{multirow}
\usepackage[pagebackref,breaklinks,colorlinks]{hyperref}

\usepackage[capitalize]{cleveref}
\crefname{section}{Sec.}{Secs.}
\Crefname{section}{Section}{Sections}
\Crefname{table}{Table}{Tables}
\crefname{table}{Tab.}{Tabs.}

\def\cvprPaperID{8416} 
\def\confName{CVPR}
\def\confYear{2022}

\begin{document}

\title{Look Closer to Supervise Better:\\ One-Shot Font Generation via Component-Based Discriminator}

\author{
Yuxin Kong,\textsuperscript{\rm 1}
Canjie Luo,\textsuperscript{\rm 1}
Weihong Ma,\textsuperscript{\rm 1 }
Qiyuan Zhu,\textsuperscript{\rm 2}
Shenggao Zhu,\textsuperscript{\rm 2}
Nicholas Yuan,\textsuperscript{\rm 2}
Lianwen Jin\footnotemark[1] \textsuperscript{\rm 1 \rm 3}\\
\textsuperscript{\rm 1}School of Electronic and Information Engineering, South China University of Technology. \\
\textsuperscript{\rm 2}Huawei Cloud AI.
\textsuperscript{\rm 3}Peng Cheng Laboratory, Shenzhen, Guangdong, China. \\
{\tt\small \{kongyxscut, canjie.luo, scutmaweihong, lianwen.jin, nicholas.jing.yuan\}@gmail.com} \\ 
{\tt\small \{zhuqiyuan2, zhushenggao\}@huawei.com}
}
\maketitle
\thispagestyle{empty}
\renewcommand{\thefootnote}{\fnsymbol{footnote}} 
\footnotetext[1]{Corresponding author.} 

\begin{abstract}
Automatic font generation remains a challenging research issue due to the large amounts of characters with complicated structures. Typically, only a few samples can serve as the style/content reference (termed few-shot learning), which further increases the difficulty to preserve local style patterns or detailed glyph structures. We investigate the drawbacks of previous studies and find that a coarse-grained discriminator is insufficient for supervising a font generator. To this end, we propose a novel Component-Aware Module (CAM), which supervises the generator to decouple content and style at a more fine-grained level, \textit{i.e.}, the component level. Different from previous studies struggling to increase the complexity of generators, we aim to perform more effective supervision for a relatively simple generator to achieve its full potential, which is a brand new perspective for font generation. The whole framework achieves remarkable results by coupling component-level supervision with adversarial learning, hence we call it Component-Guided GAN, shortly CG-GAN. Extensive experiments show that our approach outperforms state-of-the-art one-shot font generation methods.
Furthermore, it can be applied to handwritten word synthesis and scene text image editing, suggesting the generalization of our approach.

\end{abstract}

\section{Introduction}
\label{sec:intro}
\begin{figure}[t]
  \centering
    \includegraphics[width=0.9\linewidth]{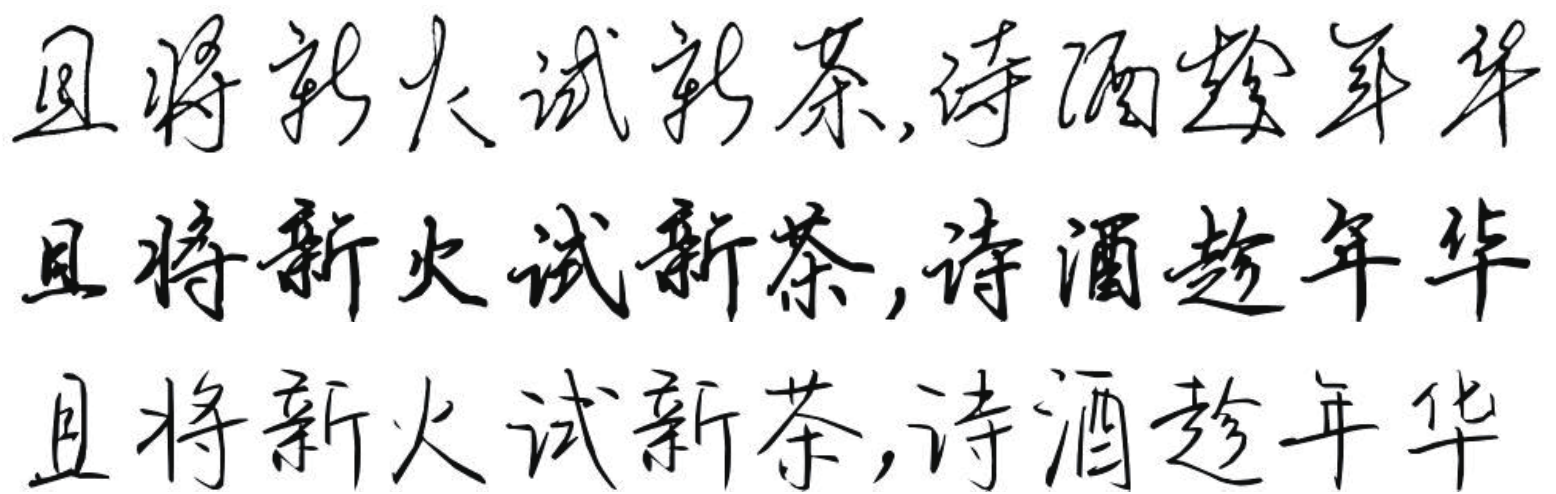}
    \caption{A same text string presented in three different font styles.}
    \label{fig_vis_intro}
\vspace{-2em}
\end{figure}

To better tackle the few-shot font generation issue, we rethink the following two questions: 1) What determines people's judgment on font styles? 2) How do people learn to write a new character/glyph in the correct structure?
To answer the first question intuitively, we present a text string in three different font styles in ~\cref{fig_vis_intro}.
Since their overall architectures are similar, we naturally pay more attention to local details, including endpoint shapes, corner sharpness, stroke thickness, joined-up writing pattern, \textit{etc.}, which appear at a more local level, \textit{i.e.}, the components of a character.
Although the components cannot present some font style properties, such as the inclination and aspect ratio, we argue that the components determine the font style to a greater extent than the whole character shape.
As for the second question, one strong assumption is that when people learn a complicated glyph, they first learn the components that form the character.
Intuitively, if all the components in a glyph are properly written, we can obtain the glyph correctly.
Drawing inspiration from the above observations, an intuitive method for few-shot font generation is to utilize the component information that is largely correlated with the font style properties and glyph structures.

Few-shot font generation (FFG) has received considerable research interest in recent years due to its critical applications~\cite{cha2020dmfont, park2020lffont, Rewrite, wu2020calligan, zhang2018EMD, zi-2-zi}.
An ideal FFG system can greatly reduce the burden of the time-consuming and labor-intensive font design process, particularly for those language systems with a massive number of glyphs, \eg, Chinese with more than 25,000 glyphs.
Another application is to create a cross-lingual font library, \eg, Chinese to Korean, given the fact that Adobe and Google take years to create the Source Han Sans Font, a universal font style that supports Chinese, Korean, and Japanese simultaneously. 

Recently, several attempts have been made to few-shot font generation; however, they all have certain limitations and further improvements are therefore required.  
For instance, ~\cite{Rewrite} learns to map the source font style to a fixed target font style, and thus has to be retrained for another new style.
``zi2zi"~\cite{zi-2-zi} learns multiple font styles by adding a pre-defined style category embedding, but still cannot generalize to unseen font styles.
One notable approach is EMD~\cite{zhang2018EMD}, which is generalizable to unseen styles by disentangling the style and content representation, but the result is not promising due to its flaw in loss function design.

Lately, several methods utilize the idea of compositionality. Still, they have significant drawbacks.
For instance, CalliGAN~\cite{wu2020calligan} generates glyph images conditioned on the learned embeddings of the component labels and style labels, hence cannot generalize to either unseen styles or unseen components.
DM-Font~\cite{cha2020dmfont} employs a dual-memory architecture for font generation.
However, it requires a reference set containing all the components to extract the stored information, which is unacceptable for the FFG scenario.
LF-Font~\cite{park2020lffont} can be extended to unseen styles conditioned on the component-wise style features.
However, its visual quality decreases significantly in the one-shot generation scenario.
Although these component-based algorithms advance by successfully encoding the diverse local styles, they explicitly depend on the component category inputs to extract style features, and thus the capability of cross-lingual font generation is entirely beyond their reach.
Meanwhile, the above methods~\cite{cha2020dmfont, park2020lffont, Rewrite, wu2020calligan, zhang2018EMD, zi-2-zi} have a common limitation, that is, they require large amounts of paired data for pixel-level strong supervision.
Although DG-Font~\cite{xie2021dgfont} achieves unsupervised font generation, the generated glyphs often contain characteristic artifacts. Overall, the performance of the state-of-the-arts is still unsatisfactory.

In this paper, we propose a novel component-guided generative network, namely CG-GAN, which might provide a new perspective for few-shot font generation.
The proposed method is inspired by two human behavior: 1) people naturally pay more attention to component parts when distinguishing font styles, and 2) people learn a new glyph by first learning its components.
Such a human learning scheme is perfectly adopted in our proposed Component-Aware Module (CAM), supervising the generator at the component level for both styles and contents.
Specifically, CAM first employs an attention mechanism for component extraction, acting as a loss function to supervise whether each component is transferred properly during the generation.
Then the learned attention maps, which represent the corresponding component information, are used to conduct per-component style classification and realism discrimination.
Finally, with the multiple component-level discriminative outputs, CAM can feedback more fine-grained information to the generator by backpropagation, encouraging the generator to simultaneously focus on three critical aspects at the component level: style consistency, structural correctness, and image authenticity.
Therefore, the quality of the generated glyph images is significantly boosted. 
Since CAM is only performed as the component-level supervision during training, it will not bring additional compute time in inference.
In addition, paired training data are not required with our method.
Once the model is trained, our generator is capable to generalize to unseen style, unseen content, even other unseen language glyphs, \textit{i.e.}, cross-lingual font generation.

Essentially, our goal is to seek an algorithm that can effectively enhance the representational ability of the generator.
The proposed CG-GAN allows to employ style-content disentanglement at a more fined grained level, \textit{i.e.}, the component level, thus enabling to extract high-quality representations from even a single reference image.
The method of utilizing component-level supervision rather than pixel-level strong supervision is a human-like method that shows effectiveness in capturing localized style patterns and preserving detailed glyph structures.
Compared with existing component-based methods, CG-GAN has two outstanding properties:1) the performance improvement is gained by providing more effective supervision for the generator, not by struggling to increase the complexity of the generator; and 2) the generator is able to capture local style patterns without explicit dependency on the predefined component categories, showing remarkable one-shot Chinese font generation and cross-lingual font generation abilities.

Extensive experiments demonstrate that our proposed CG-GAN significantly outperforms state-of-the-arts in one-shot font generation. Furthermore, by coupling the component-level guidance with a novel framework design, CG-GAN can flexibly extend to two other different tasks: handwriting generation and scene text editing, producing stunning results that far exceed our expectations, indicating the significant potential of our proposed method.

\begin{figure*}[t]
  \setlength{\abovecaptionskip}{-1em}
  \begin{center}
  \includegraphics[width=0.99\textwidth]{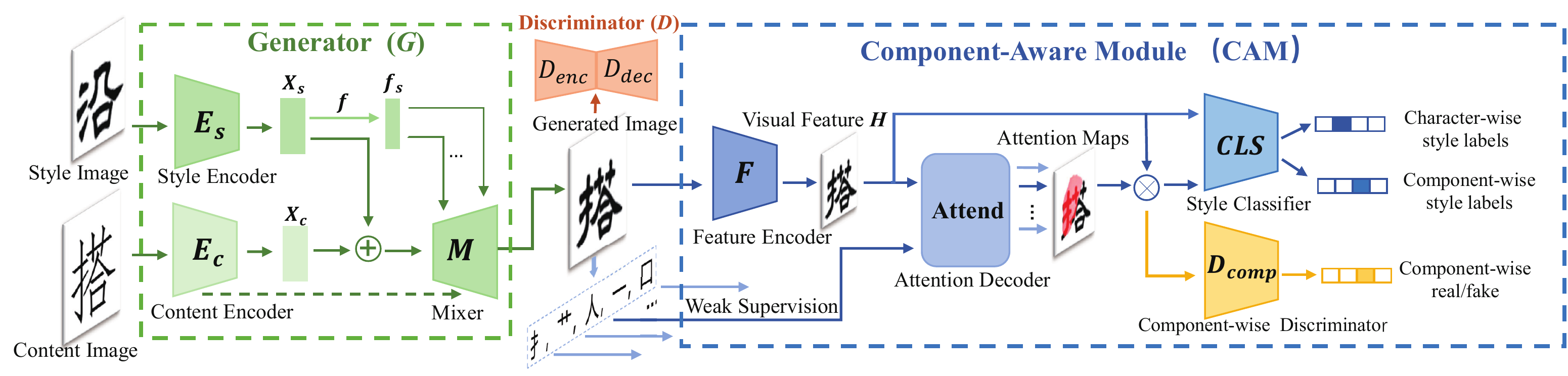}
  \end{center}
     \caption{Overview of the proposed method. }
  \label{fig_overall_pipeline}
\vspace{-1.2em}
  \end{figure*}
\section{Related works}
\label{sec:related}
\subsection{Image-to-Image Translation}
Image-to-image (I2I) translation aims to translate an input image in the source domain into a corresponding output image in the target domain.
Pix2pix~\cite{isola2017pix2pix} is the first general framework for the I2I translation task, which is in supervised learning and built upon the conditional adversarial network~\cite{mirza2014conditional}.
However, paired training data are unavailable for many scenarios.
Therefore, several methods are proposed to tackle the unpaired setting.
UNIT~\cite{liu2017UNIT} is an extension of CoGAN~\cite{liu2016cogan}, which learns the joint distribution across domains with the combination of generative adversarial networks~\cite{goodfellow2014generative} and variational autoencoders~\cite{kingma2013auto}.
One notable work is CycleGAN~\cite{zhu2017cyclegan}, which tackles the unsupervised image translation problem by introducing the cycle consistency loss.
Concurrent to CycleGAN~\cite{zhu2017cyclegan}, DiscoGAN~\cite{kim2017discoGAN} and DualGAN~\cite{yi2017dualgan} also utilize the cycle consistent constraint to implement unpair I2I translation.
The above approaches are limited to translating images between two classes.
Later,~\cite{anoosheh2018combogan, choi2018stargan, hui2018unsupervised} are proposed to achieve multi-class unsupervised I2I translation, capable of translating images among multiple seen classes.
FUNIT~\cite{liu2019funit} further extends its generalization ability to unseen classes by learning to encode the content images and class images respectively.

\subsection{Few-shot Font Generation}
Few-shot font generation (FFG) aims to create a complete font library in the required style given only a few reference images.
Several methods~\cite{chang2018chinese,lyu2017auto,Rewrite,zi-2-zi} regard the FFG task as an I2I translation problem as both tasks learn a mapping from the source domain to the target domain. For instance,
``Rewrite"~\cite{Rewrite} is built upon the pix2pix framework and learns to transfer only one font style. 
DC-font~\cite{jiang2017dcfont} learns the transformation relationship between two fonts in deep space via the feature reconstruction network. However, all of them cannot generalize to unseen styles.
After that, EMD~\cite{zhang2018EMD} and SA-VAE~\cite{sun2017SAVAE} are proposed to separate the representations for styles and contents, and are thus generalizable to unseen styles.
However, they fail to capture local style patterns.
Later, some component-based methods~\cite{huang2020rd-gan,cha2020dmfont,park2020lffont,wu2020calligan, park2021multiple} are proposed. For example,
RD-GAN~\cite{huang2020rd-gan} can generate unseen glyphs in a one-shot setup by introducing a radical extraction module, but can only transfer to a fixed target font style. 
~\cite{cha2020dmfont,park2020lffont} improve the generative quality by learning component-wise style representation, where DM-Font~\cite{cha2020dmfont} introduces a dual memory structure and LF-Font~\cite{park2020lffont} utilizes a factorization strategy.
However, they suffer from estimation errors and cannot generalize to cross-lingual font generation due to the explicit dependency on the components labels.
To tackle the cross-lingual scenario, MX-Font~\cite{park2021multiple} employs a multi-headed encoder design, where each head can extract different localized features in a weak component supervision manner.
All of the above works are in supervised learning and require paired data for strong supervision.
After that, DG-Font~\cite{xie2021dgfont} achieves unsupervised learning by introducing a deformation skip connection, but the results contain artifacts.

\section{Methodology}
\label{sec:methods}
The overall architecture of our proposed CG-GAN is shown in ~\cref{fig_overall_pipeline}, which consists of a generator $G$, a Component-Aware Module (CAM) and a discriminator $D$. 
The generator $G$ aims to implement style-content disentanglement at the component level.
To this end, CAM is employed to provide component-level feedback to the generator through multi component-level discrimination outputs.
A U-Net based discriminator~\cite{schonfeld2020unet} $D$ is also employed to perform per-image and per-pixel discrimination, further enhancing the quality of the generated glyph.

\subsection{Generator}
As shown in ~\cref{fig_overall_pipeline}, the generator consists of a style encoder, a content encoder and a mixer.
Given a style image $I_{s}$ and a content image $I_{c}$, the generated image $I_{g}$ should present the font style of $I_{s}$ while maintaining the same underlying structure of $I_{c}$.
Specifically, the content encoder encodes the input content image into a style-invariant content feature map $X_{c}$.
Meanwhile, the style encoder is employed to extract the style representation at two different levels from the style reference image: a style feature map $X_{s}$ and a style latent vector $f_{s}$. Here, 
$X_{s}$ is extracted from the style reference image, and is later mapped to a style latent vector $f_{s}$ through a mapping network $f$, which is implemented using a multi-layer perception (MLP). 

Finally the mixer is employed to integrate the style and content representation and reconstruct the target image.
The style feature map $X_{s}$ and the content feature map $X_{c}$, which have the same spatial dimensions, are concatenated in the channel-wise dimension and are later fed to the mixer.
Meanwhile, the style latent vector $f_{s}$ is injected into each up-sampling block of the mixer $M$ through the AdaIN~\cite{huang2017adain} operation.
In addition, we adopt the skip-connection between the content encoder and the mixer, where the output of every down-sampling block in the content encoder is concatenated to the input of the up-sampling block that has the same resolution in the mixer.

\subsection{Component-Aware Module}
\label{subsec:cam}
Intuitively, a glyph font style and structure are closely related to the component information.
However, most existing methods~\cite{jiang2017dcfont, Rewrite, zhang2018EMD, zi-2-zi} employ pixel-level strong supervision while ignoring the critical component information.
Therefore, we introduce the Component-Aware Module (CAM), where the main idea is to make full use of the component information to better guide the font generation process.
Hence, CAM is designed to supervise the generator at the component level using the following strategies:

\textbf {Component\;extraction} A prerequisite for font generation is to preserve the detailed structure of the target glyph.
Therefore, the component extraction process aims to supervise whether the glyph structure is correctly transferred.
Since every Chinese glyph can be decomposed into a unique component set following a depth-first reading order, we treat the component extraction process as a sequential problem.
To proceed, the CNN-based feature encoder $\mathcal{F}$ extracts high-level visual features from the input image $\mathbf{x}$: $\boldsymbol{H} = \mathcal{F} (\boldsymbol{x})$, where $\boldsymbol{H}$ has a spatial dimension of $C\times H \times W$.
Thus the encoder output $\boldsymbol{H}$ is a feature vector of $L=H\times W$ elements, where each element $h_{i}$ is a $C$-dimensional vector that represents its corresponding region in the input image,
denoted by $\boldsymbol{H} = \{ h_{1}, h_{2}, ..., h_{L}\}$.

Compared with other sequential learning methods, \eg, CTC \cite{graves2006ctc} and Transformer \cite{vaswani2017transformer}, the attention mechanism is particularly well suited for our purposes due to its efficiency and ease of convergence.
Therefore, we adopt the attention-based decoder $\mathcal{A}$ to generate the component sequences, denoted by $\boldsymbol{Y} = \{ y_{1}, y_{2}, ..., y_{T}\}$, where T is the length of the component sequence.
Note that the length of the component sequence is variable.
The decoder predicts the output sequence one symbol at a time until an end-of-sequence token $'EOS'$ is predicted.
At every time step t, output $y_{t}$ is,
\begin{equation}
  y_{t} = Softmax(W_{o}x_{t}+b_{o}), \label{equ-2}
\end{equation}
where $x_{t}$ is the output vector at time step t.
We update $x_{t}$ alongs with the hidden state $s_{t}$ using a gated recurrent unit (GRU) as:
\begin{equation}
  (x_{t}, s_{t}) = \mathrm{GRU}(({g_{t}}, {y}_{prev}), s_{t-1}), \label{equ-3}
\end{equation}
where $({g_{t}}, {y}_{prev})$ is the concatenation of the glimpse vectors $g_{t}$ and the embedding vetors of the previous output ${y}_{t-1}$;  $g_{t}$ is calculated through the attention mechanism as follows:
\begin{align}
    &{y}_{prev} = \mathrm{Embedding} ({y}_{t-1}),  \label{equ-4}\\
    &e_{t} = \tanh(s_{t-1}W_{s}+y_{prev}W_{y}+b), \label{equ-5}\\
    &{\alpha_{t,i}} = {\exp(e_{t,i})}  /  {\sum_{j=1}^L(\exp(e_{t,j}))}, \label{equ-6}\\
    &g_{t}  = \sum_{i=1}^L(\alpha_{t,i} h_{i}), \label{equ-7}
  \end{align}
where $W_{o},b_{o},W_{s},W_{y}$ and $b$ are trainable parameters; $h_{i}$ denotes the $i$-th feature vector in the input feature map $\boldsymbol{H}$. 

\begin{figure}[t]
  \setlength{\abovecaptionskip}{0em}
  \centering
   \includegraphics[width=0.85\linewidth]{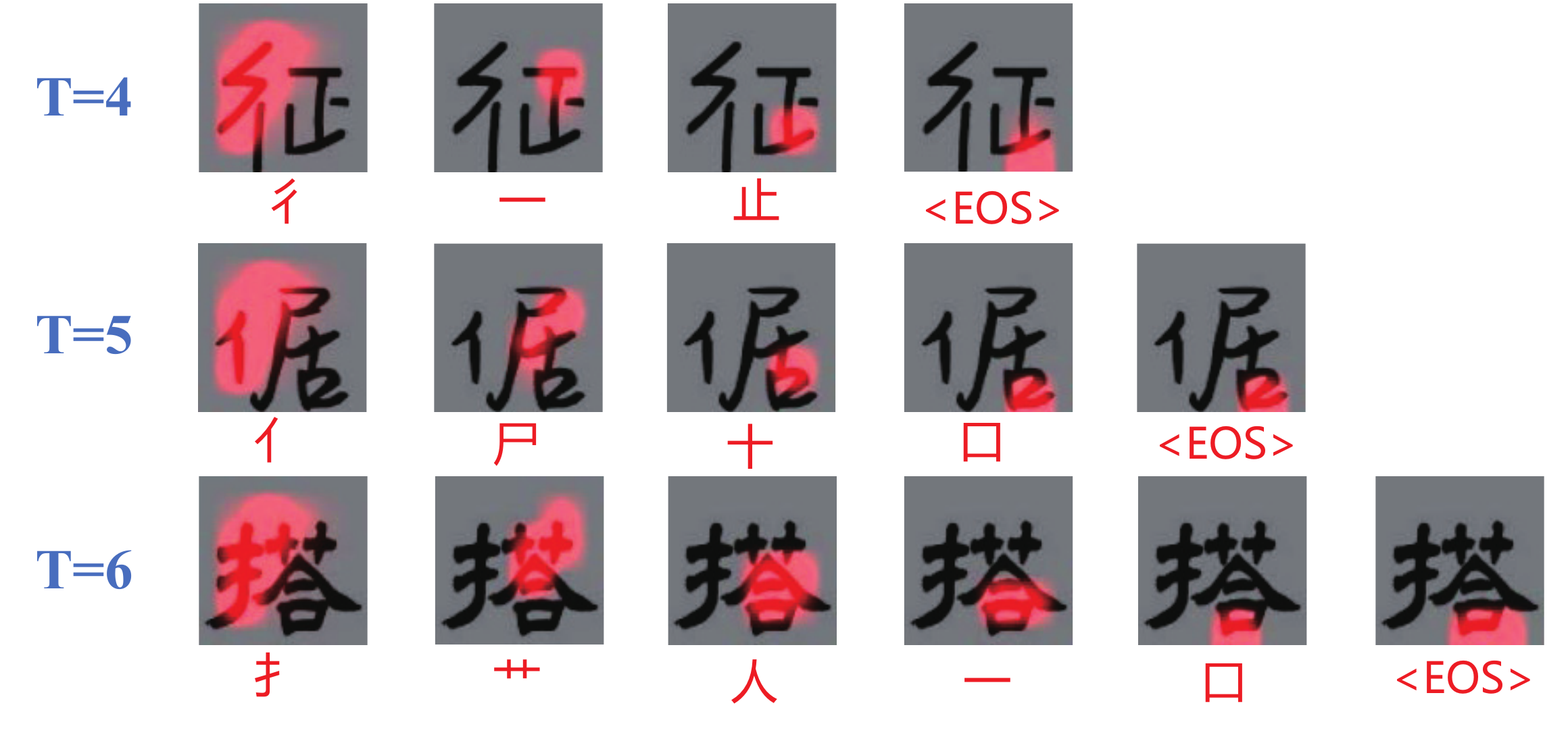}
   \caption{Visualization of attention maps on different lengths of component sequences. Symbols below the images are the predicted components.}
   \label{fig:visual_CAM}
  \vspace{-1em}
\end{figure}

Using only component labels as weak supervision, the attention-based decoder is able to localize every component by minimizing the structure retention loss. Note that $\mathcal{F}$ and $\mathcal{A}$ are only optimized with the real samples $I_{s}$, denote as:
\begin{equation}
   \mathcal{L}^{CAM}_{strc} = \mathbb{E}_{I_{s} \in P_s}\Big[-\sum^{T}_{i=1}\widehat{y}_{t} \log(\mathcal{A}(\mathcal{F}(I_{s})))_{t}\Big],
\end{equation}
where $\widehat{y}_{t}$ denotes the corresponding ground-truth component label at time step $t$. As shown in \cref{fig:visual_CAM}, at every time step t, the decoder is able to focus on the corresponding component region.
Thus, if the component prediction of the generated glyph goes wrong, the generator $G$ is penalized for an incorrect structure transfer, denote as:
\begin{equation}
  \mathcal{L}^{G}_{strc} = \mathbb{E}_{I_{s} \in P_s,I_{c} \in P_c}\Big[-\sum^{T}_{i=1}\widehat{y}_{t} \log(\mathcal{A}(\mathcal{F}(G(I_{s},I_{c}))))_{t}\Big].
\end{equation}

In this manner, $G$ is supervised to generate glyph structure at the component level, devoted to preserving every single component correctly. 
Unlike most existing methods that only extract a global content representation, which often leads to incomplete structures, we employ $\mathcal{L}^{G}_{strc}$ to supervise the content encoder $E_{c}$ at the component level, guiding $E_{c}$ to actively decompose the content representation from $I_{c}$ at the component level. 
Such a learning scheme is more capable of handling the enormous Chinese categories, as well as preserving the complicated structures. 

\textbf{Multi component-level discrimination} We further introduce a style classifier $CLS(\cdot)$ and a discriminator $D_{comp}$ to conduct component-level discrimination.
Intuitively, people naturally pay more attention to the local parts/components but less to the entire shape when distinguishing different font styles. Therefore, we conduct the style classification and realism discrimination by utilizing the attention maps $\boldsymbol{A} = \{\alpha_{1}, \alpha_{2}, ..., \alpha_{T}\},\; \boldsymbol{\alpha_{t}} \in \mathbb{R}^{H\times W}$ as the label of component regions. To effectively guide the generation process, we conduct multi-component-level discrimination for each input image simultaneously. 

\subsection{Loss Function}

\textbf{Adversarial loss} The generator $G$ aims to synthesize a realistic image that is indistinguishable from real samples. Hence, we adopt a U-Net based discriminator \cite{schonfeld2020unet}, where the encoder part $D_{enc}$ and the decoder part $D_{dec}$ perform per-image and per-pixel discrimination, respectively. Therefore, the generator now has to fool both $D_{enc}$ and $D_{dec}$ via the adversarial loss:
\begin{equation}
    \mathcal{L}_{adv} = \mathcal{L}^{enc}_{adv} + \lambda_{dec} \mathcal{L}^{dec}_{adv},
\end{equation}
\begin{equation}
   \mathcal{L}^{enc}_{adv} = \mathbb{E}_{I_{s} \in P_s, I_{c} \in P_c} [ \log D_{enc}(I_{s}) +\log (1-D_{enc}(G(I_{s}, I_{c})))],
\end{equation}
\begin{equation}
\begin{aligned}
   \mathcal{L}^{dec}_{adv} = \mathbb{E}_{I_{s} \in P_s, I_{c} \in P_c} \Big[\sum_{i,j} \log [D_{dec}(I_{s})]_{i.j}\\
+ \sum_{i,j} \log (1-[D_{dec}(G(I_{s}, I_{c}))]_{i,j})\Big],
\end{aligned}
\end{equation}
where $[D_{dec}(\cdot)]_{i,j}$ denotes the discrimination output at position$(i,j)$. We set $\lambda_{dec}$ to 0.1 in our experiments. 

\textbf{Style matching loss} In addition to using structure retention loss to supervise the structure correctness (~\cref{subsec:cam}), the generated image should also maintain both global and local style coherence.
To this end, the style classifier $CLS(\cdot)$ performs the style classification over the whole input image to ensure global style coherence, along with performing this classification on a per-component basis to estimate the local style consistency.
Therefore, the style matching loss is computed by considering the above two aspects simultaneously. Given the 2D-attention map ${\alpha_{t}}$ at time step t and the corresponding style label $w$ of the reference style image $I_{s}$, the style matching loss is defined as:
\begin{equation}
\begin{aligned}
   \mathcal{L}^{CAM}_{sty} = \mathbb{E}_{I_{s} \in P_s}\Big[-w \log(CLS(\mathcal{F}(I_{s}))
   \\
   -\sum^{T}_{t=1}w \log(CLS(\alpha_{t} \otimes \mathcal{F}(I_{s})) \Big].
\end{aligned}
\end{equation} 
Here, $\otimes$ refers to an element-wise multiplication. Note that $CLS(\cdot)$ is only optimized with real samples $I_{s}$, thus it can guide the generator to synthesize images with a highly similar font style $w$ to the reference image $I_{s}$.
Correspondingly, the generator is optimized by minimizing:
\begin{equation}
\begin{aligned}
    \mathcal{L}^{G}_{sty} = \mathbb{E}_{I_{s} \in P_s,I_{c} \in P_c}\Big[-w \log(CLS(\mathcal{F}(G(I_{s},I{c}))
  \\
  -\sum^{T}_{t=1}w \log(CLS(\alpha_{t} \otimes \mathcal{F}(G(I_{s},I_{c})))) \Big].
\end{aligned}
\end{equation}
Essentially, the $\mathcal{L}^{G}_{sty}$ enforces the style encoder $E_{s}$ to disentangle the style representation at the component level, thus enabling the $E_{s}$ to capture diverse local styles while maintaining global style coherence. Notably, the $\mathcal{L}^{G}_{sty}$ results in a more powerful style encoder $E_{s}$, which can accurately encode local style patterns from any reference style sample $I_{s}$ without accessing the corresponding component labels.

\textbf{Component realism loss} Furthermore, a discriminator $D_{comp}$ is additionally employed to classify each component patch into being real or fake, further supervising the visual verisimilitude of $I_{gen}$ at the component level, denoted as:
\begin{equation}
\begin{aligned}
  \mathcal{L}_{comp} = \mathbb{E}_{I_{s} \in P_s, I_{c} \in P_c} \Big[ \log D_{comp}(\mathcal{F}(I_{s}))
\\
+\sum^{T}_{i=1}\log (1-D_{comp}(\alpha_{t} \otimes \mathcal{F}(G(I_{s},I_{c}))) \Big],
\end{aligned}
\end{equation}
encouraging the generator to pay more attention to local realism of the generated glyph images.


\textbf{Identity loss} We additionally adopt the identity loss to 
guarantee the identity mapping in the generator $G$: the generator $G$ is able to reconstruct the style reference image $I_{s}$ when $I_{s}$ is also provided as the content input, \textit{i.e.},
\begin{equation}
    \mathcal{L}_{idt} = \mathbb{E}_{I_{s} \in P_s}\left\|I_{s} - G(I_{s},I_{s})\right\|_1.
\end{equation}
This identity loss stabilizes the training process to a certain extent, as it avoids an excessive style transfer.
\textbf{Content loss} We adopt a content loss to guarantee that the extracted content representation $X_{c}$ is style-invariant, denoted as:
\begin{equation}
\mathcal{L}_{cnt} = \mathbb{E}_{I_{s} \in P_s, I_{c} \in P_c}\left\|X_{c} - E_{c}(G(I_{s}, I_{c}))\right\|_1.
\end{equation}

\textbf{Full objective}  Finally, the discriminator $D$, the Component-Aware Module CAM and the generator $G$ of our proposed CG-GAN are optimized respectively as
\begin{equation}
    \mathcal{L}_{D} = -\mathcal{L}_{adv},   
\mathcal{L}_{CAM} = -\mathcal{L}_{comp} + \mathcal{L}^{CAM}_{strc} + \mathcal{L}^{CAM}_{sty},
\end{equation}
\begin{equation}
  \mathcal{L}_{G} = \mathcal{L}_{adv} + \mathcal{L}_{comp} + \mathcal{L}^{G}_{strc} + \mathcal{L}^{G}_{sty} + \mathcal{L}_{idt} + \lambda_{cnt} \mathcal{L}_{cnt}.
\end{equation}
The entire framework is trained from scratch in an end-to-end manner. We set $\lambda_{cnt}$ to 10 in our experiments. 
\begin{figure*}
  \centering
  \begin{subfigure}{1\linewidth}
    \centering
    \includegraphics[width=0.85\textwidth]{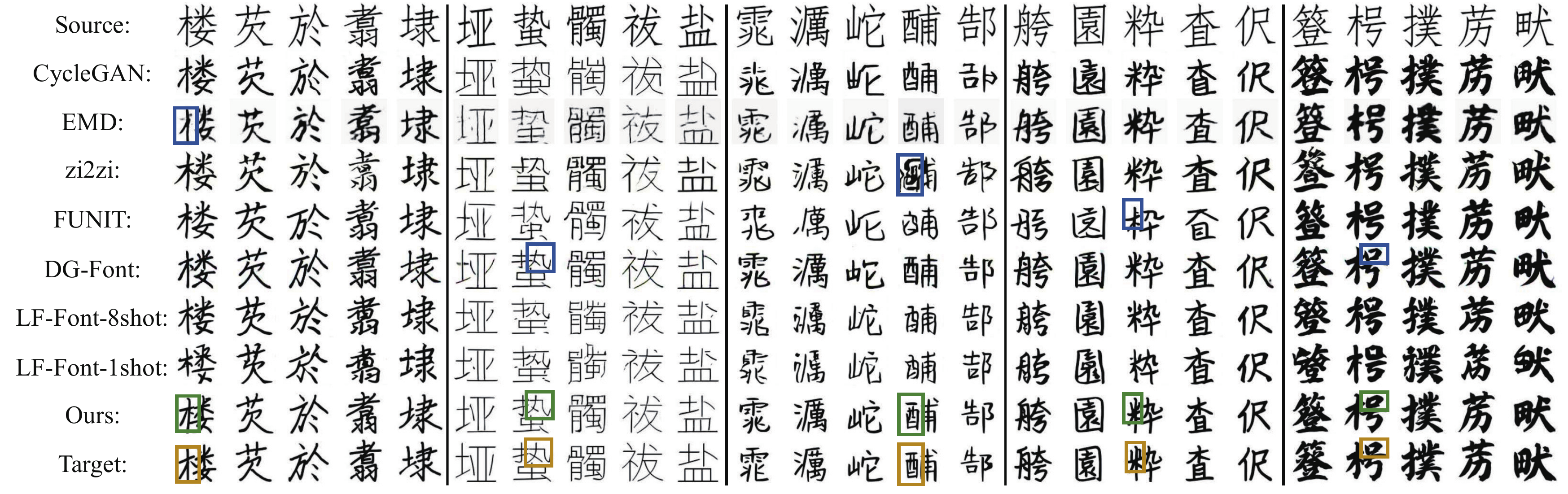}\label{fig:4a}
    \caption{Seen styles and unseen contents.}
    \label{fig:short-a}
  \end{subfigure}
  \hfill
  \begin{subfigure}{1\linewidth}
    \centering
    \includegraphics[width=0.85\textwidth]{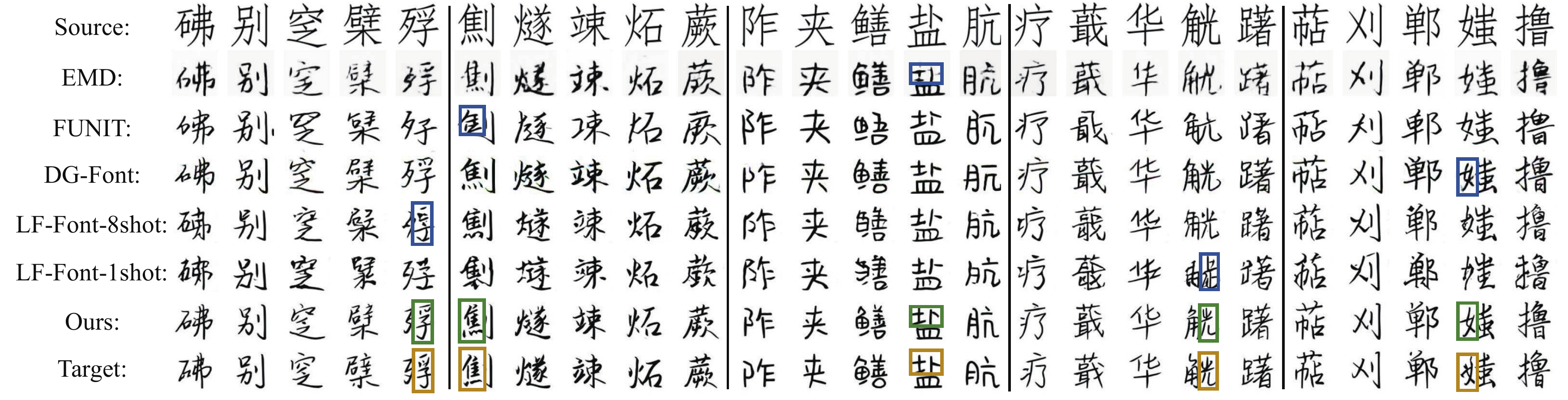}\label{fig:4b}
    \caption{Unseen styles and unseen contents.}
    \label{fig:short-b}
  \end{subfigure}
  \begin{subfigure}{1\linewidth}
    \centering
    \includegraphics[width=0.85\textwidth]{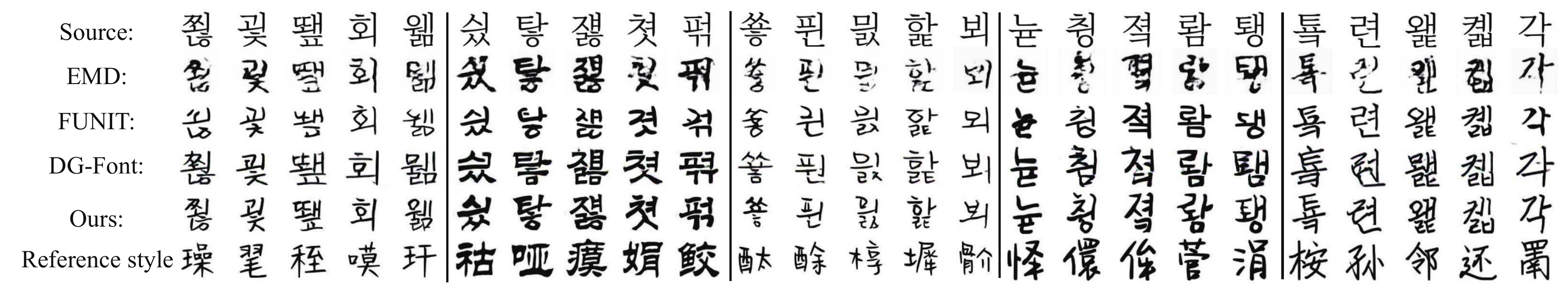}\label{fig:4c}
    \caption{Cross lingual font generation.}
    \label{fig:short-c}
  \end{subfigure}
  \caption{Comparisons with the state-of-the-art methods for font generation.}
  \label{fig_vis_font}
 \vspace{-1em}
\end{figure*}

\section{Experiments}
\label{sec:exp}

\subsection{Chinese font generation}
\textbf{Datasets} To evaluate our method with the Chinese font generation task, we collect a dataset containing 423 fonts.
We randomly select 399 fonts as the training set (\textit{i.e.} seen fonts), where each font contains 800 Chinese characters (\textit{i.e.} seen characters) that can be decomposed by 385 components.
We evaluate the one-shot Chinese font generation ability on two test sets: One is the 399 seen fonts with 150 unseen characters per font.
The other is the remaining 24 unseen fonts with 200 unseen characters per font.  
We additionally evaluate the generalization ability to the unseen language glyphs by using a Korean glyph test set consisting of the 24 unseen fonts with 200 Korean characters per font. 

\textbf{Comparison with state-of-the-art methods} 
We compared our model with six state-of-the-art methods, including four few-shot Chinese font generation methods (zi2zi~\cite{zi-2-zi}, EMD~\cite{zhang2018EMD}, LF-font \cite{park2020lffont}, DG-Font~\cite{xie2021dgfont}), and two unsupervised image-to-image translation methods (CycleGAN~\cite{zhu2017cyclegan}, FUNIT~\cite{liu2019funit}). For a fair comparison, we use the font Song as the source font which is a common setting in the font generation task \cite{zhang2018EMD, xie2021dgfont,park2020lffont}.
Since Cycle-GAN can only learn the mapping from one domain to another at a time, we train a total of 399 CycleGAN models.
LF-Font exhibits a low visual quality in inference if only one reference sample is provided. Hence, we evaluate its performance in an eight-shot setup (its original setting) and a one-shot setup respectively. All models are trained from scratch using their official code.

\textbf{Evaluation metrics} We use several metrics for quantitative evaluation.
First, SSIM and RMSE are employed to measure whether pixel-level details can be preserved, and a higher SSIM and a lower RMSE represent less image distortion of the generated images.
Second, LPIPS \cite{zhang2018lpips} is adopted to quantify the perceptual similarity; where a lower LPIPS denotes the generated image is more in line with human visual perception.
Third, FID \cite{heusel2017FID} is employed to measure whether the model can match the target data domain distribution. A lower FID represents a higher quality and variety of the generated images.
Finally, a user preference study is conducted to quantify the subjective quality of output images.
We randomly select 30 seen fonts and 20 unseen fonts from the two Chinese glyph test sets.
At each time, the participants are shown the reference style image along with $n$ generated samples generated by the $n$ different methods, and asked to pick the best result.
In total, we have collected 2,400 responses in two scenarios of seen styles and unseen styles respectively, contributed by 48 participants.

\textbf{Quantitative comparison} The quantitative results are shown in Table~ \ref{table1}.
Except for the LF-Font-eight-shot, all the reported results are tested in a one-shot setting.
As shown in Table~\ref{table1}, CG-GAN achieves the best performance on all the evaluation metrics for both seen styles and unseen styles.
Particularly, CG-GAN outperforms previous state-of-the-arts with significant gaps in both perceptual-level metrics and human visual preference, \eg, achieves 8.92 lower FID in seen styles and 10.87 lower FID in unseen styles than the second-best LF-Font-8-shot, and gain more than 60\% of the visual choice under both scenarios.
Notably, with just one shot, CG-GAN still outperforms the second-best LF-Font-eight-shot, which further demonstrates the powerful generation ability of our proposed method.

\begin{table}[t]
  \setlength{\abovecaptionskip}{0em}
  \footnotesize
  \caption{\textbf{Quantitative evaluation on the whole dataset}. We evaluate methods on seen styles and unseen contents, unseen styles and unseen contents. The bold number indicates the best.} 
  \begin{center}
  \setlength{\tabcolsep}{4pt}{
  \begin{tabular}{llllll}
  \hline
  Methods & SSIM$\uparrow$  & RMSE$\downarrow$  & LPIPS$\downarrow$  & FID$\downarrow$  & \makecell[c]{User \\ preference (\%) }  \\
  \hline
  \multicolumn{6}{c}{\textbf{Seen} styles and \textbf{Unseen} contents } \\
  \hline
  cycleGAN \cite{zhu2017cyclegan} & 0.7092 & 0.0247 & 0.3010 & 64.85  & \makecell[c]{4.48} \\
  FUNIT \cite{liu2019funit} & 0.7269 & 0.0244 & 0.2720 & 57.72  & \makecell[c]{5.24} \\
  zi2zi \cite{zi-2-zi} & 0.7666 & 0.0216 & 0.2268 & 59.79  & \makecell[c]{3.43} \\
  EMD \cite{zhang2018EMD}  & 0.7519 & 0.0213 & 0.2536 & 61.29  & \makecell[c]{0.10} \\
  DG-Font \cite{xie2021dgfont}  & 0.7697 & \textbf{0.0212} & 0.2079 & 41.56  & \makecell[c]{8.19} \\
  8-LF-Font \cite{park2020lffont} & 0.7535 & 0.0223 & 0.2227 & 15.46  & \makecell[c]{13.81} \\
  1-LF-Font \cite{park2020lffont} & 0.7427 & 0.0232 & 0.2499 & 19.36  & \makecell[c]{-} \\
  CG-GAN (ours) & \textbf{0.7703} & \textbf{0.0212} & \textbf{0.1919} & \textbf{6.54}  & \makecell[c]{\textbf{64.76}} \\
  \hline
  \multicolumn{6}{c}{\textbf{Unseen} styles and \textbf{Unseen} contents } \\
  \hline
  FUNIT \cite{liu2019funit} & 0.7074 & 0.0249 & 0.2892 & 64.68  & \makecell[c]{3.90} \\
  EMD \cite{zhang2018EMD}  &  0.7373 & 0.0219 & 0.2620 & 84.84  & \makecell[c]{0.19} \\
  DG-Font \cite{xie2021dgfont}  & 0.7553 & 0.0221 & 0.2195 & 55.73  & \makecell[c]{13.52} \\
  8-LF-Font \cite{park2020lffont}  & 0.7419 & 0.0225 & 0.2295 & 28.81  & \makecell[c]{8.00} \\
  1-LF-Font \cite{park2020lffont} & 0.7310 & 0.0232 & 0.2529 & 33.49  & \makecell[c]{-} \\
  CG-GAN (ours) & \textbf{0.7568} & \textbf{0.0218} & \textbf{0.2058} & \textbf{17.94}  & \makecell[c]{\textbf{74.38}} \\
  \hline
  \end{tabular}}
\end{center}
\label{table1}
\vspace{-3em}
\end{table}

\textbf{Qualitative comparison} In ~\cref{fig_vis_font} (a) and (b), we provide a visual comparison of seen styles and unseen styles, which intuitively explains the significant gaps of CG-GAN in the user preference study.
For these two challenging scenarios, our method generates glyph images of higher quality than state-of-the-arts, particularly better satisfying both style consistency and structure correctness.
CycleGAN and FUNIT often produce results in an incomplete structure.
EMD often produces severe blur and an unclear background. zi2zi loses some detailed structure if the target glyph is complex.
LF-Font suffers from a significant decrease in visual quality if only one reference image is provided.
DG-Font generates glyphs containing characteristic artifacts, which can be observed in the highlighted region. 
As shown in~\cref{fig_vis_font} (c), we further test the generalization ability to unseen components, \textit{i.e.}, cross-lingual font generation.
Owing to a stronger representation capability, our model shows superior cross-lingual FFG performance.

\subsection{Handwriting generation}
By coupling component-level supervision with a novel framework design, CG-GAN can be directly applied to handwriting generation task without any adjustment. To evaluate this, we conduct experiments on IAM handwriting dataset \cite{marti2002iam}. IAM dataset consists of 9,862 text lines with 62,857 handwritten words, contributed by 500 different writers. In our experiments, only the training and validate sets are used for model training, and the test set is kept apart for evaluation. For a fair comparison, we evaluate our methods with the state-of-the-art handwriting generation methods under the following two scenarios:
   
\textbf{Writer-relevant handwriting generation} Following previous studies~\cite{bhunia2021handwriting, kang2020ganwriting}, we first evaluate the writer-relevant scenario, where FID is calculated for each writer between its corresponding generated samples and real samples, and finally average the sum of FID of all writers. Thus the final FID scores evaluate the generative quality and simultaneously, the style imitation capability.
We use HWT \cite{bhunia2021handwriting} and GANwriting \cite{kang2020ganwriting} as our baselines, which can synthesize images with referenced style.
Shortly, HWT is a transformer-based method that can synthesize arbitrary-length texts.
GANwriting can generate short word images with no more than ten letters. We evaluate the competing methods in four different settings: IV-S, IV-U, OOV-S, OOV-U, respectively. 

As shown in Table~\ref{table2}, our proposed CG-GAN shows comparable performance with the state-of-the-arts. We surpass the second-best HWT under three settings of IV-S, OOV-S, and OOV-U.
Particularly for the most challenging one, where both words and styles have never been seen during training(OOV-U), CG-GAN still achieves around 1.0 lower FID than the second-best HWT. Note that both HWT and GANwriting use 15 style reference images for training, and their proposed results are tested in a 15-shot setting, whereas our method is trained and tested under only a 1-shot setting.

\begin{figure}[t]
  \centering
   \includegraphics[width=0.85\linewidth]{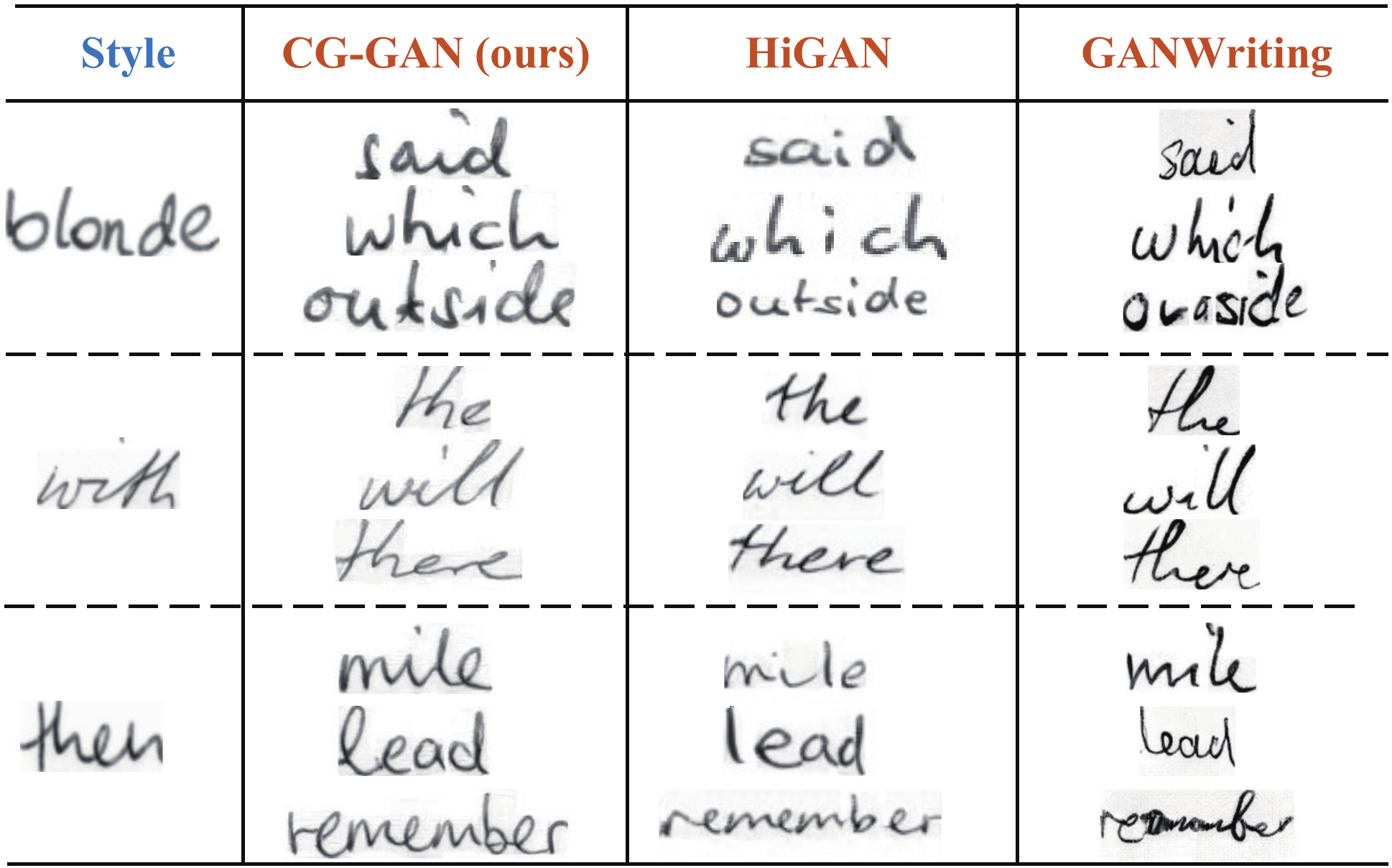}
   \caption{Visual comparison for synthesizing handwritten words.}
   \label{fig_vis_eng}
 \vspace{-1em}
\end{figure}

\textbf{Writer-irrelevant handwriting generation} We further evaluate the writer-irrelevant scenario, where writer identity is ignored when calculating the FID score.
Under this scenario, we use HWT \cite{bhunia2021handwriting}, ScrabbleGAN \cite{fogel2020scrabblegan} and HiGAN \cite{gan2021higan} as our baselines.
Briefly, ScrabbleGAN can synthesize long texts with random styles but cannot imitate referenced styles. HiGAN can synthesis an arbitrary-length text either with a random or referenced style. As shown in Table~\ref{table3}, our method achieves comparable performance with the state-of-the-arts. A visual comparison is shown in \cref{fig_vis_eng}.
\begin{table}[t]
  \centering
  \caption{\textbf{Writer-relevant handwriting generation quality comparison.} All four settings: In-Vocabulary words and seen style (IV-S), In-Vocabulary words and unseen style (IV-U), Out-of-vocabulary words and seen styles (OOV-S), Out-of-vocabulary words and unseen styles (OOV-U).}
  \begin{adjustbox}{width=0.42\textwidth}
  \begin{tabular}{c c c c c}
    \toprule
            & IV-S & IV-U  & OOV-S & OOV-U \\
      \midrule
      GAN writing \cite{kang2020ganwriting} & 120.07  & 124.30  & 125.87 & 130.68 \\
      \midrule
      HWT \cite{bhunia2021handwriting} & 106.97  & \textbf{108.84}  & 109.45 & 114.10 \\
      \midrule
      CG-GAN (ours)  & \textbf{102.18}  & 110.07 & \textbf{104.81} & \textbf{113.01} \\
      \bottomrule
  \end{tabular}
\end{adjustbox}
\label{table2}
\vspace{-0.5em}
\end{table}
\begin{table}[t]
  \centering
  \caption{\textbf{Writer-irrelevant handwriting generation quality comparison.} Writer identity is ignored when calculating FID.}
  \begin{adjustbox}{width=0.42\textwidth}
  \begin{tabular}{c c c c c}
    \toprule
    Method & ScrabbleGAN \cite{fogel2020scrabblegan} & HiGAN \cite{gan2021higan} & HWT \cite{bhunia2021handwriting} & CG-GAN (ours) \\
    \midrule
    FID & 23.78 & \textbf{17.28} & 19.40 & 19.03 \\
    \bottomrule
  \end{tabular}
  \end{adjustbox}
  \label{table3}
  \vspace{-1.2em}
\end{table}

\subsection{Ablation study}
We perform multiple ablation studies to evaluate the effectiveness of our proposed CAM on the one-shot Chinese font generation task.
The results are tested on the unseen styles test dataset.

\textbf{Effectiveness of the Component-level supervision}
We compare our component-level supervision with commonly used pixel-level supervision and character-level supervision.
Pixel-level supervision is performed by removing the CAM module and replacing the component-level objectives with the L1 loss.
Note that pixel-level supervision is trained using paired data which uses the same reference style images as our unpaired data settings. 
Character-level supervision is implemented by replacing the component labels with the character labels.
In this manner, the loss supervision is conducted at the character level.
As shown in Table~\ref{table5}, We can see that the quantitative results obviously improve in terms of SSIM, RMSE, LPIPS and FID, which demonstrate the effectiveness of our proposed component-level supervision.

\textbf{Effectiveness of the Component-Aware Module}
We further analyze the influence of each component-level supervision provided by CAM.
First, we build a baseline, remove the CAM module and replace it with a style classifier at the image level, and thus the baseline no longer contains any component-level supervision.
Next, we successively add different parts of the multi-component-level supervision and analyze their impact, including structure retention loss, style matching loss and component realism loss.
The results are reported in Table~\ref{table6}, we can observe that all of our proposed component-level objective functions are essential, the addition of each objective can make a further improvement on both visual quality and quantitative results.

\section{Extension}
\begin{figure}[h]
\setlength{\abovecaptionskip}{0.2em}
  \centering
   \includegraphics[width=0.99\linewidth]{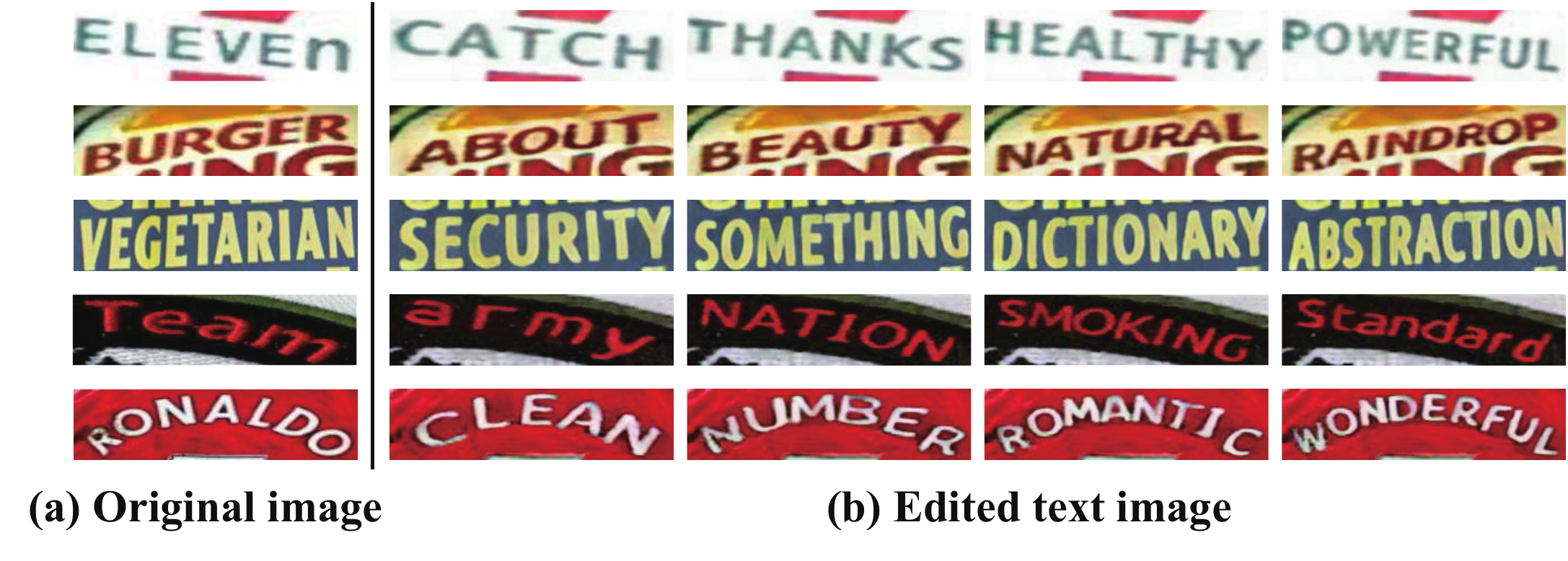}
   \caption{Visualization of scene text editing results. (a) and (b) are original text images and text edited images with different text contents and lengths.}
   \label{fig_vis_edit}
 \vspace{-1.5em}
\end{figure}

Our framework can be further extended to the scene text editing (STE) task, which is challenging due to large variations in font style, text shape and background.
Existing STE methods~\cite{wu2019editing,yang2020swaptext} generally approach this task in two stages: first rendering the target textual content to obtain the text-modified foreground, and erasing the original text to obtain the text-erased background, finally fusion the two to obtain the desired target image. However, these two-stage methods do not generalize well to real-world scene text images due to strong mutual interference of the background and foreground. By contrast, our framework abandons the inefficient multi-stage rendering and alleviates the intervention problem with the help of component-level supervision.
As shown in~\cref{fig_vis_edit}, our framework generates quite promising results that exceed our expectations, showing the impressive potential of our proposed approach. The implementation details are shown in the Appendix A.
\begin{table}[t]
  \centering
  \caption{Effectiveness of the Component-level supervision.}
  \begin{adjustbox}{width=0.42\textwidth}
  \begin{tabular}{c c c c c}
    \toprule
            Method & SSIM$\uparrow$ & RMSE$\downarrow$  & LPIPS$\downarrow$ & FID$\downarrow$ \\
      \midrule
      pixel-level  & 0.7479  & 0.0223  & 0.2298 & 51.44 \\
      \midrule
      character-level  & 0.7529  & 0.0223  & 0.2142 & 33.21 \\
      \midrule
      component-level  & \textbf{0.7568}  & \textbf{0.0218}  & \textbf{0.2058} & \textbf{17.94} \\
      \bottomrule
    \end{tabular}
  \end{adjustbox}
  \label{table5}
  \vspace{-0.5em}
\end{table}
\begin{table}[t]
  \centering
  \caption{Effectiveness of the Component-Aware Module.}
  \begin{adjustbox}{width=0.42\textwidth}
  \begin{tabular}{c c c c c}
      \toprule
            & SSIM$\uparrow$ & RMSE$\downarrow$  & LPIPS$\downarrow$ & FID$\downarrow$ \\
      \midrule
      baseline  & 0.7517  & 0.0225  & 0.2251 & 49.09 \\
      \midrule
      +$\mathcal{L}_{strc}$  & 0.7487  & 0.0227  & 0.2138 & 22.33 \\
      \midrule
      +$\mathcal{L}_{strc}$+$\mathcal{L}_{sty}$  & 0.7532  & \textbf{0.0216}  & 0.2084 & 18.67 \\
      \midrule
      +$\mathcal{L}_{strc}$+$\mathcal{L}_{sty}$+$\mathcal{L}_{dcomp}$  & \textbf{0.7568}  & 0.0218 & \textbf{0.2058} & \textbf{17.94} \\
      \bottomrule
  \end{tabular}
  \end{adjustbox}
  \label{table6}
  \vspace{-1em}
\end{table}

\section{Conclusion}
In this paper, we propose a simple yet effective CG-GAN for one-shot font generation. Specifically, we introduce a CAM to supervise the generator. The CAM decouples the style and content at a more fine-grained level, \textit{i.e.}, the component level, to guide the generator to achieve more promising representation ability. Furthermore, to our best knowledge, CG-GAN is the first FFG method that can be potentially extended to both handwriting word generation and scene text editing, showing its generalization ability.

\section*{Acknowledgement}
This research is supported in part by NSFC (Grant No.: 61936003) and GD-NSF (No.2017A030312006, No.2021A1515011870).

{
\bibliographystyle{ieee_fullname}
\bibliography{arxiv}
}
\appendix

\setcounter{page}{1}

\twocolumn[
\centering
\Large
\textbf{Look Closer to Supervise Better:\\ One-Shot Font Generation via Component-Based Discriminator} \\
\vspace{0.5em}Supplementary Material \\
\vspace{1.0em}
] 

\appendix

\section{Implementation}
\subsection{Training details}

The model is optimized using Adam with the settings of $\beta_{1}$=0.5 and $\beta_{2}$=0.999. All modules are trained from scratch with a learning rate of 0.0001. We initialize the weights of convolutional and linear layers with a Guassian distribution $\mathcal{N}$(0,0.02). The batch size is set to 16 in all experiments. Our method is implemented in PyTorch and all experiments are conducted on a single NVIDIA 1080Ti GPU.

\textbf{Chinese font generation} All the images are resized to $128\times128$ pixels. The learning rate is initially set to 0.0001 and linearly decreased to zero after 40 epochs.

\textbf{Handwriting generation} The images are resized to a height of 64 pixels, and the width is calculated with the original aspect ratio (up to 384 pixels). We keep the learning rate as 0.0001 for the first 15 epochs and linearly decay the rate to zero over the next 30 epochs.

\textbf{Scene text editing} For the scene text editing task, the model is trained with synthetic data and evaluated on real-world scene text image data. Specifically, we generate 1.4M synthetic data ($I_{s},I^{'}_{s}$) with the synthesizing engine SynthTiGER~\cite{yim2021synthtiger}, where $I_{s}$ and $I^{'}_{s}$ have different textual content ($T, T^{'}$) respectively but other image properties such as background, font, \textit{etc}. remain the same. In the training process, we use $I_{s}$ as the style reference input and meanwhile, render the textual content $T^{'}$ into a content reference image, which is used as the content reference input. Since the scene text image lacks a style label, we set the style retention loss to zero and add the perceptual loss~\cite{johnson2016perceptual} and the spatially-correlative loss~\cite{zheng2021spatially} on the basis of the original training objectives. The test set is sampled from regular and irregular scene text datasets, including IIIT5k~\cite{mishra2012it5k}, SVT~\cite{wang2011svt}, IC03~\cite{icdar03}, IC13~\cite{karatzas2013icdar}, SVT-P~\cite{phan2013svtp}, CUTE80~\cite{risnumawan2014cute80} and IC15~\cite{karatzas2015ic15}, with a total of 9,350 real-world scene text images. All the images are resized to $64\times256$ pixels and the model is trained for 20 epochs with a learning rate of 0.0001.

\subsection{Network architectures} \textbf{Generator architecture}  Our generator is built upon the ResNet architecture of ~\cite{zhang2019self}, and is further extended with our proposed changes. The original generator of ~\cite{zhang2019self} is an encoder-decoder architecture. In order to obtain the font generator, we adopt the original encoder architecture as our style encoder and content encoder, while using the original decoder architecture as our mixer, with a channel multiplier $ch=64$. Specifically, the style encoder and the content encoder have the same architecture, consisting of five ResNet down-sampling blocks with a total down-sampling rate of 32. In the mixer, the encoded features are upsampled via five ResNet up-sampling blocks until the original image resolution is reached. To produce the $3\times H \times W$ output image, an InstanceNorm-ReLU-conv2d block with output channel 3 is additionally appended as the last layer of the mixer. We remove the self-attention layer in all ResNet blocks and add AdaIN operation as the normalization layer in every up-sampling block of the mixer. 

\textbf{CAM architecture} The proposed CAM aims to supervise the generator at the component level. The detailed architecture of the CAM is shown in Table~\ref{architecture1}.

\textbf{Discriminator architecture} For the discriminator networks, we adopt a U-Net based discriminator~\cite{schonfeld2020unet}. Specifically, We adopt the U-Net discriminator architecture of the $128\times 128$ resolution with a channel multiplier $ch=16$.

\section{Additional qualitative results}
In this section, We present more qualitative results and ablation study results to better validate the effectiveness of our proposed method. 

\subsection{One-shot font generation} 
\begin{figure*}
  \centering
   \includegraphics[width=0.95\linewidth]{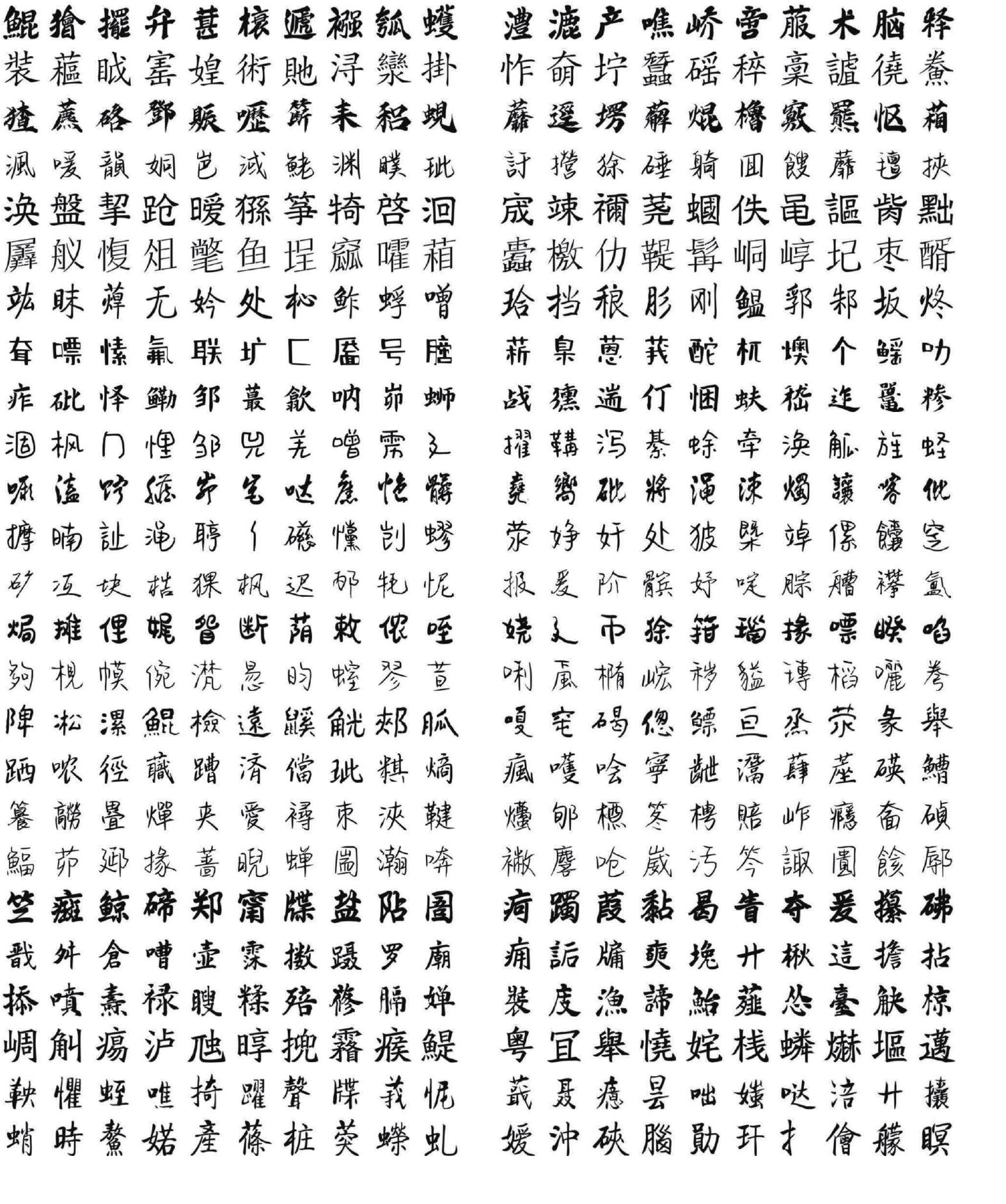}
   \caption{Seen styles and unseen contents in Chinese one-shot font generation.}
   \label{fig_seen}
\end{figure*}

\begin{figure*}
\centering
 \includegraphics[width=0.8\linewidth]{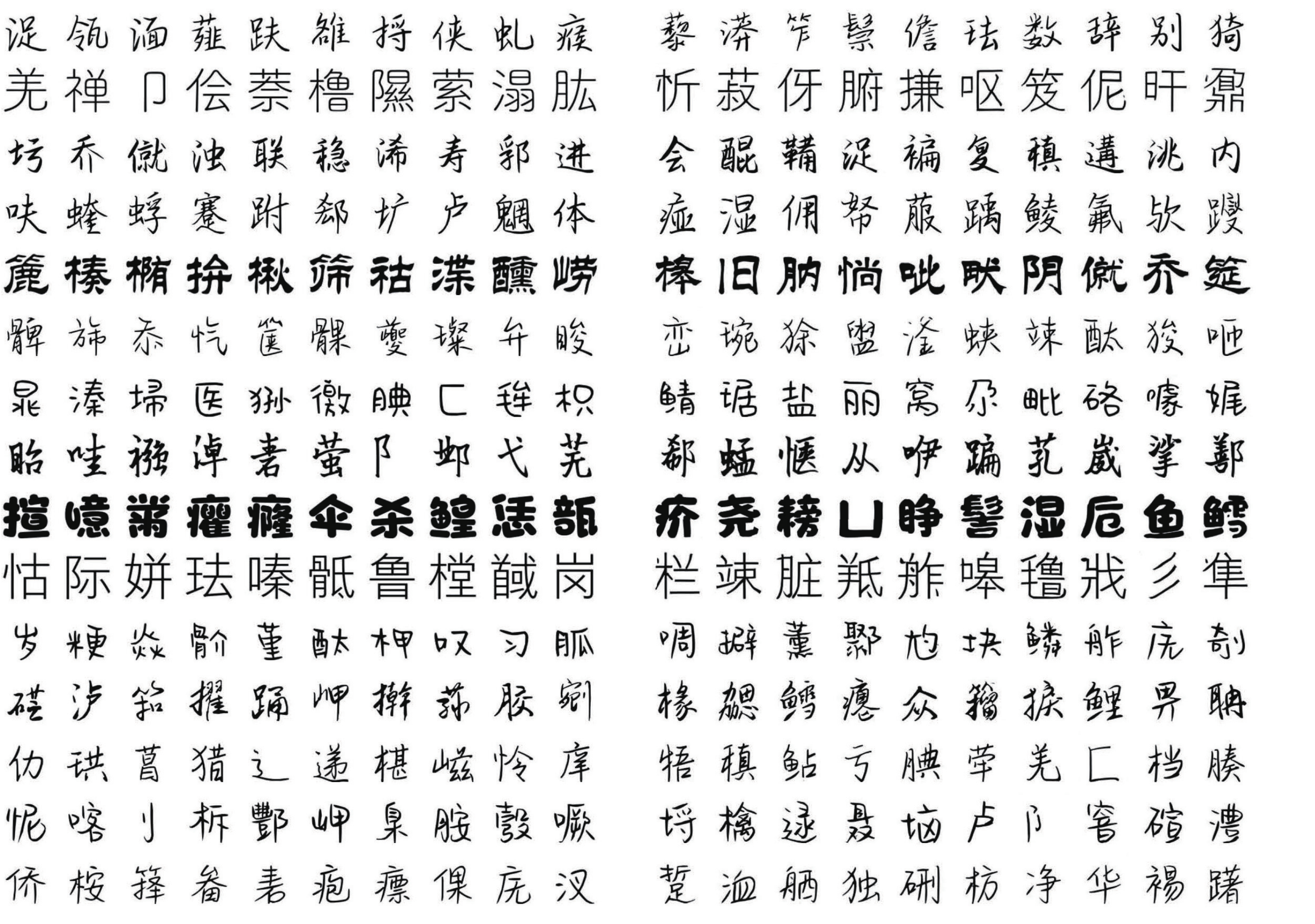}
 \caption{Unseen styles and unseen contents in Chinese ont-shot font generation.}
 \label{fig_unseen}
\end{figure*}

\begin{figure*}
\centering
 \includegraphics[width=0.8\linewidth]{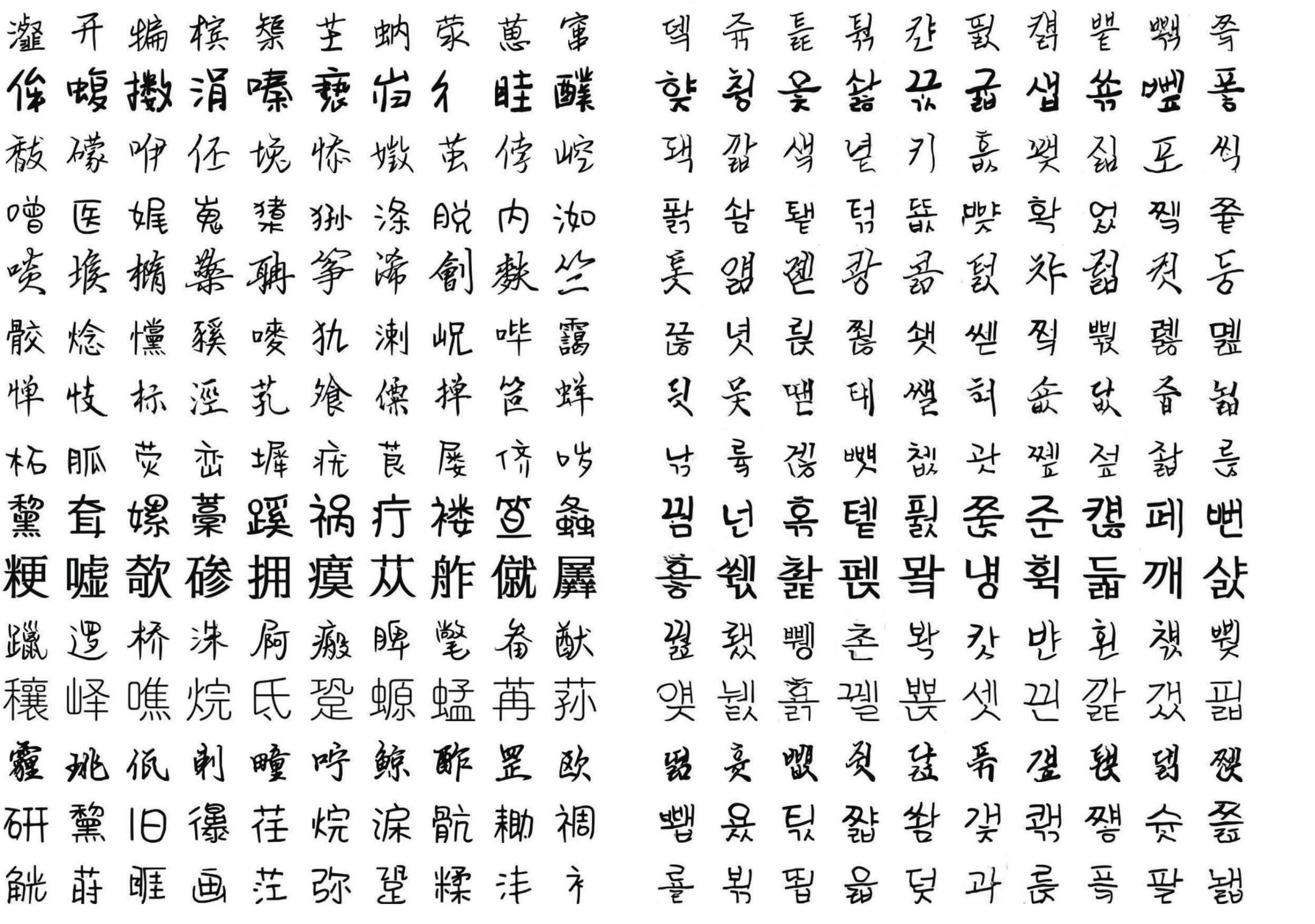}
 \caption{Cross lingual font generation.}
 \label{fig_cross}
\end{figure*}

In ~\cref{fig_seen} and ~\cref{fig_unseen}, we present more generated samples in two scenarios: seen styles and unseen styles, respectively. Specifically, we randomly select 30 seen fonts and 20 unseen fonts from the two Chinese glyph test sets, and randomly sample 10 unseen target glyphs for each font to carry out the qualitative evaluation. Note that all the generated glyphs are tested in a one-shot setting, with one single style reference image provided. The results show that CG-GAN can generate high-quality glyph images in both scenarios, suggesting the superior one-shot font generation ability. ~\cref{fig_cross} shows that our model is able to extend to cross-lingual font generation. The model is trained on Chinese fonts but is able to generate a complete Korean font library in inference.

\begin{table*}[t]
 \caption{\textbf{CAM architecture}. BN denotes the batch normalization, and IN denotes the Instance normalization}
  \scalebox{0.9}{
  \begin{tabular}{cccccccc}
  \hline
                                   & Operation                              & Kernel size & Resample & Padding & Feature maps & Normalization & Nonlinearity \\ \hline
  \multirow{5}{*}{Feature encoder}   & Convolution                            & 7           & MaxPool  & 3       & 96           & BN            & PReLU         \\
                                   & Convolution                            & 3           & MaxPool  & 1       & 128          & BN            & PReLU         \\
                                   & Convolution                            & 3           & MaxPool & 1       & 160          & BN            & PReLU         \\
                                   & Convolution                            & 3           & -  & 1       & 256          & BN            & PReLU         \\
                                   & Convolution                            & 3           & MaxPool & 1       & 256          & BN            & PReLU         \\
                                   
                                   \hline

Attention decoder & &  & \makecell[c]{256 hidden units, \\ 256 GRU units } & &  &        \\
                                   \hline
                                   
  \multirow{4}{*}{Style classifier}           & Convolution & 3           & MaxPool & 1       & 256          & IN         & PReLU         \\
                                   & Convolution                            & 3           & - & 1       & 512          & IN         & PReLU         \\
                                   & Convolution                            & 3           & MaxPool & 1       & 512           & IN         & PReLU         \\
                                   & Convolution                            & 3           & - & 1       & n styles            & -             & -       \\ \hline
  \multirow{4}{*}{Component-wise discriminator}           & Convolution & 3           & MaxPool & 1       & 128          & IN         & PReLU         \\
                                   & Convolution                            & 3           & MaxPool & 1       & 64          & IN         & PReLU         \\
                                   & Convolution                            & 3           & - & 1       & 16           & IN         & PReLU         \\
                                   & Convolution                            & 3           & - & 1       & 1            & -             & -       \\ \hline
  \end{tabular}}
 
  \label{architecture1}
  \end{table*}
  
\subsection{Latent space interpolations} 
\begin{figure*}
  \centering
    \includegraphics[width=0.95\linewidth]{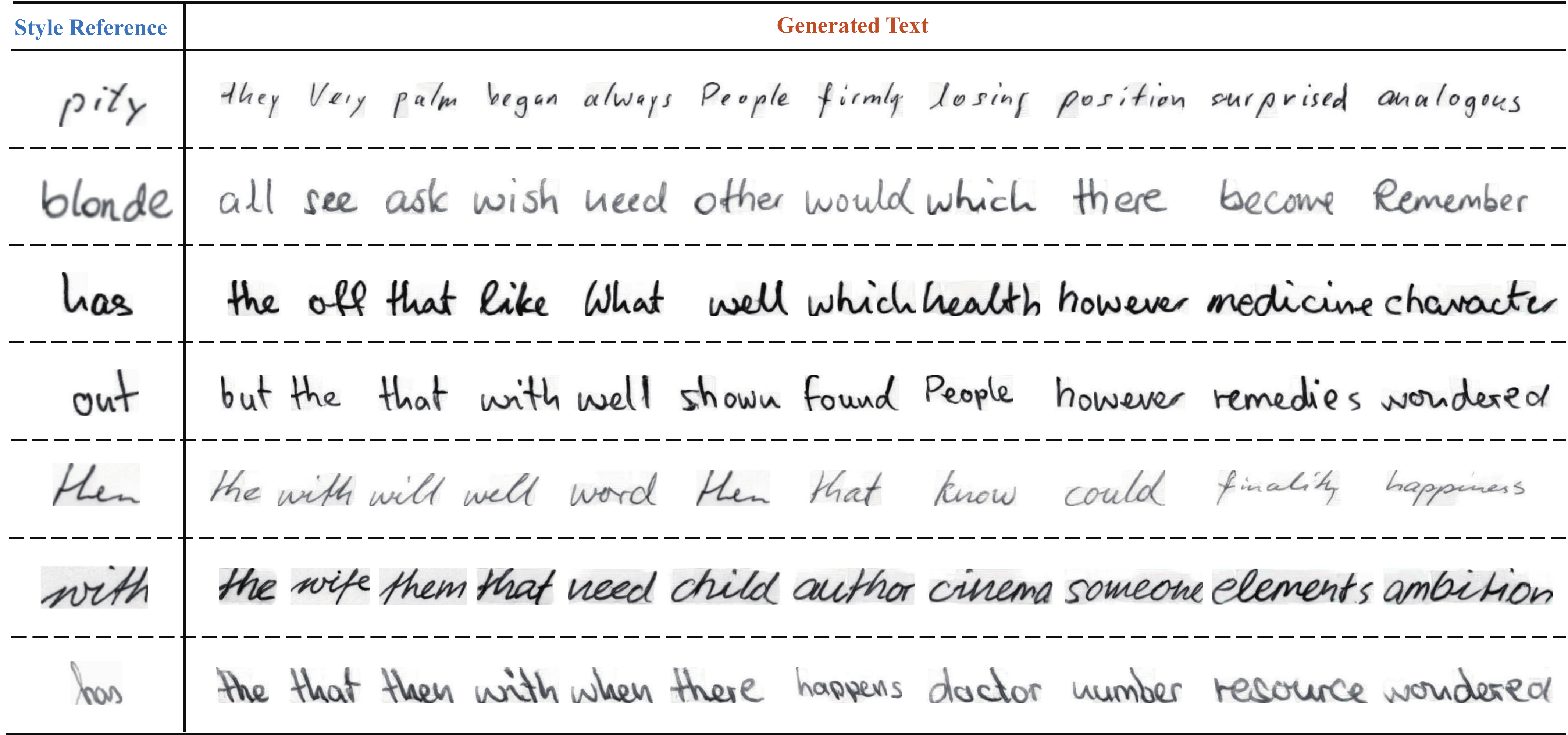}
    \caption{Visual comparison for synthesizing handwritten words.}
    \label{fig_handwritten}
\end{figure*}

\begin{figure*}
  \centering
    \includegraphics[width=0.95\linewidth]{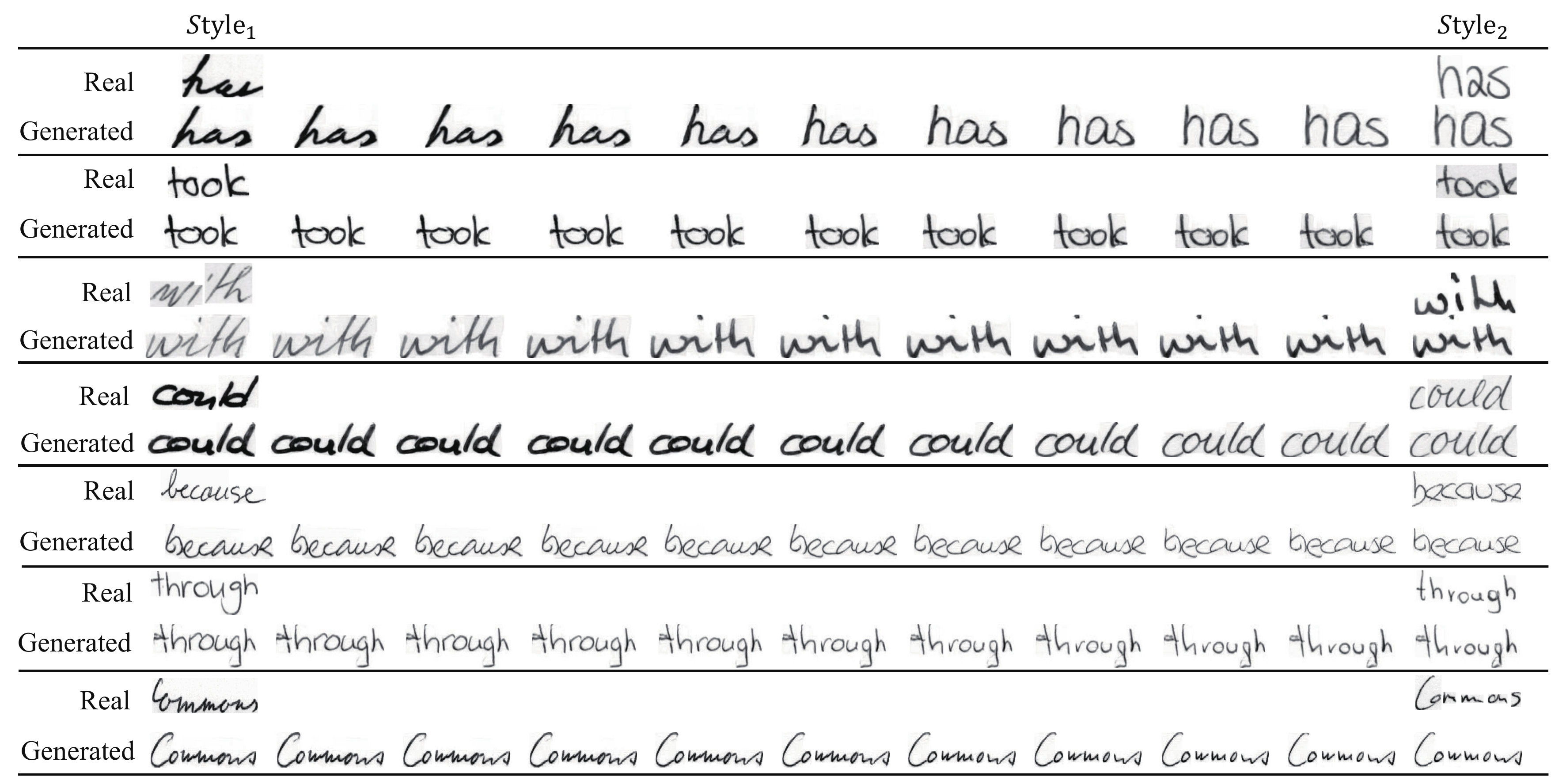}
    \caption{Style interpolation between two different styles.}
    \label{fig_interpolate}
\end{figure*}

In ~\cref{fig_interpolate}, we perform a linear style interpolation between two random styles on the IAM dataset. We can observe that the generated image can smoothly change from one style to another, while strictly preserving its textual content. The results indicate that CG-GAN can generalize in the style latent space rather than memorizing some specific style patterns. Besides, we present some synthetic word images with various calligraphic styles in ~\cref{fig_handwritten}, where each row presents diverse generated samples in the same calligraphic style. 

\subsection{Scene text editing} 
\begin{figure*}[t]
  \centering
    \includegraphics[width=0.95\linewidth]{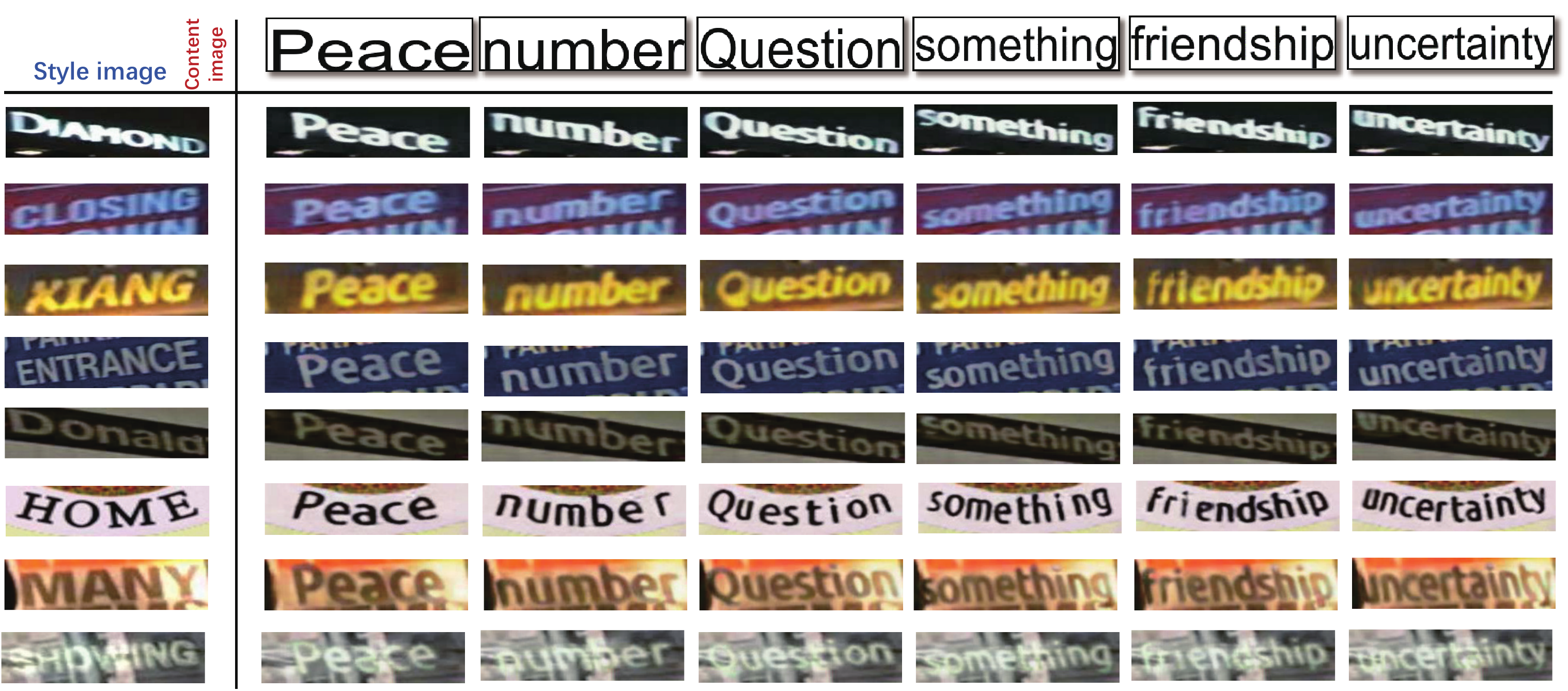}
    \caption{Additional scene text editing results.}
    \label{fig_edit}
\end{figure*}

In ~\cref{fig_edit}, we present more scene text editing results. As we can observe, our model can robustly edit textual contents with different lengths, and achieve promising results even in challenging cases, such as complex backgrounds or slanted or curved texts.

\subsection{Additional ablation results} 
\textbf{Influence of the style latent vector} In this part, we trained a variant where the AdaIN operation including the style latent vector $f_{s}$ is removed. Results are shown in Table~\ref{table-2}. It is noted that there is only a slight drop in performance, indicating that the style latent vector $f_{s}$ is not that necessary.
Such results partly reflect our primary purpose, that is, the performance improvement is mainly gained by providing more effective supervision for the generator, not by struggling to increase the complexity of the generator. 

\textbf{Influence of the U-net Discriminator} We further investigate the influence of the U-net architecture of the discriminator. Specifically, we trained a variant where the U-net architecture of the discriminator is removed, only the encoder part $D_{enc}$ is preserved. For font generation and handwriting generation tasks, we set the channel multiplier $ch$ of the discriminator to 16 and 64, respectively.
As shown in Table~\ref{table-3}, the performance of the variant is comparable to our current approach on the font generation task, which still outperforms all the other baselines in all metrics. 
And the variant is also competitive on handwriting generation task, as shown in Table~\ref{table-4}. The results indicate that decoder part $D_{dec}$ has no significant effect on the performance. This may be due to the simple background of the dataset, which contains a lot of pixels with values (255,255,255), thus the pixel-level discrimination performed by $D_{dec}$ may not be so effective.

\begin{table}[h]
  \footnotesize
  \centering
  \caption{The impact of the style latent vector on the Chinese font generation task.}
  \begin{adjustbox}{width=0.42\textwidth}
  \begin{tabular}{c c c c c}
    \toprule
      Method & SSIM$\uparrow$ & RMSE$\downarrow$  & LPIPS$\downarrow$ & FID$\downarrow$ \\
      \midrule
      CG-GAN (ours)  & \textbf{0.7568}  & \textbf{0.0218} & \textbf{0.2058} & \textbf{17.94} \\
      w/o style latent & 0.7549 & 0.0225 & 0.2193 & 18.73 \\
      \bottomrule
  \end{tabular}
\end{adjustbox}
\label{table-2}
\vspace{-1.2em}
\end{table}

\begin{table}[h]
  \setlength{\abovecaptionskip}{-0em}
  \setlength{\belowcaptionskip}{-0.4em}
  \footnotesize
  \caption{The impact of the U-net architecture of the discriminator on the Chinese font generation task.}
  \begin{center}
    \begin{adjustbox}{width=0.42\textwidth}
  \setlength{\tabcolsep}{4pt}{
  \begin{tabular}{ccccc}
  \toprule
  Method & SSIM$\uparrow$ & RMSE$\downarrow$  & LPIPS$\downarrow$ & FID$\downarrow$ \\
  \hline
  \multicolumn{5}{c}{Seen styles and Unseen contents } \\
  \hline
  CG-GAN (ours) & 0.7703 & 0.0212 & 0.1919 & \textbf{6.54} \\
  w/o $D_{dec}$      & \textbf{0.7795} & \textbf{0.0207} & \textbf{0.1821} & 7.14 \\
  \hline
  \multicolumn{5}{c}{Unseen styles and Unseen contents } \\
  \hline
  CG-GAN (ours) & 0.7568 & 0.0218 & 0.2058 & \textbf{17.94} \\
  w/o $D_{dec}$      & \textbf{0.7603} & \textbf{0.0214} & \textbf{0.1967} & 19.07 \\
  \bottomrule
  \end{tabular}}
\end{adjustbox}
\end{center}
\label{table-3}
\vspace{-1.2em}
\end{table}
\begin{table}[h]
    \centering
    \caption{The impact of the U-net architecture of the discriminator on the writer-relevant handwriting generation task.}
    \begin{adjustbox}{width=0.42\textwidth}
    \begin{tabular}{c c c c c}
      \toprule
              & IV-S & IV-U  & OOV-S & OOV-U \\
      \midrule
        CG-GAN (ours)  & 102.18  & \textbf{110.07} & 104.81 & 113.01 \\
        w/o $D_{dec}$      & \textbf{101.48} & 111.29 & \textbf{102.67} & \textbf{112.77} \\
      \bottomrule
    \end{tabular}
  \end{adjustbox}
  \label{table-4}
  \vspace{-1.2em}
  \end{table}

\newpage
\ 
\newpage
\ 
\newpage
\ 
\newpage
\ 
\newpage

   \quad

\end{document}